\documentclass[sigconf, nonacm]{acmart}

\renewcommand\footnotetextcopyrightpermission[1]{}
\AtBeginDocument{%
  }

\usepackage{amsmath}
\usepackage{algorithmic}
\usepackage{graphicx}
\usepackage{textcomp}
\usepackage{xcolor}
\usepackage{hyperref}
\usepackage{booktabs}
\usepackage{subcaption}
\usepackage[normalem]{ulem}
\usepackage{enumitem}
\usepackage[ruled,vlined,linesnumbered]{algorithm2e}
\usepackage{courier}
\usepackage{tikz}
\usetikzlibrary{shapes.geometric}
\usepackage{tabularx}

\SetEndCharOfAlgoLine{}

\begin{document}

\title{UniSpike: Accelerating Spiking Neural Networks on Neuromorphic Systems via Eliminating Address Redundancy}

\author{Qinghui Xing, Zhuo Chen, Xin Du\textsuperscript{\dag}, Ouwen Jin, Ming Zhang\\Pan Lv, Ying Li, Shuiguang Deng, Gang Pan\textsuperscript{\dag}}
\thanks{\textsuperscript{\dag}Xin Du and Gang Pan are corresponding authors.}
\affiliation{%
  \institution{Zhejiang University, Hangzhou, China \\ \{xindu, gpan\}@zju.edu.cn}
  \city{}
  \country{}
}

\renewcommand{\shortauthors}{Xing et al.}

\begin{abstract}
\par Many-core neuromorphic systems accelerate Spiking Neural Networks (SNNs), yet their packet-based spike communication can spend substantial traffic and energy repeatedly transmitting destination addresses. This overhead is amplified by the small payload of spike packets: in representative workloads, duplicate address transmissions account for up to $49\%$ of the total traffic.
\par This paper presents \textit{UniSpike}, a hardware-software co-design that removes address redundancy by aggregating spikes destined for the same core into compact packets. UniSpike combines destination-centric spike scheduling, lightweight runtime packet assembly hardware, and destination-aware SNN partitioning. Across diverse SNN workloads, UniSpike reduces traffic by 1.93$\times$ on average, delivering 1.77$\times$ speedup and 1.50$\times$ energy efficiency improvement over state-of-the-art designs.

\end{abstract}

\maketitle

\section{Introduction}
\par Neuromorphic systems are specialized accelerators for Spiking Neural Networks (SNNs), a bio-inspired computing model gaining increasing attention \cite{SNN-survey2}. Modern systems such as IBM TrueNorth \cite{debole2019truenorth}, SpiNNaker \cite{hoppner2021spinnaker2}, and Intel Loihi \cite{davies2018loihi} use many-core architectures in which cores exchange spike events through a packet-based Network-on-Chip (NoC). This inter-core spike traffic is a major bottleneck for latency and energy \cite{communication-bottleneck2}.

\par Existing optimizations mainly target two forms of redundancy: \textit{payload redundancy}, where one firing neuron sends the same information to multiple destination cores and is often handled by multicast \cite{multicast-path-TC14, HamDP, HamMP, HamK}; and \textit{data redundancy}, where repeated or compressible packet content is reduced by compression \cite{das2008performance, jin2008adaptive,deb2021flitzip}. We identify a different source of inefficiency: \textit{address redundancy}. Many spike packets generated by different neurons often target the same destination core, repeatedly carrying identical destination addresses. Because spike packets contain only minimal payloads, these repeated addresses form a large fraction of NoC traffic and are not removed by multicast or packet compression.

\par To address this issue, we propose \textit{UniSpike}, a hardware-software co-designed communication scheme. UniSpike first schedules spike transmission from the destination perspective so spikes sharing a destination can be aggregated. It then uses lightweight hardware, including a TS Manager and bitmap-based Packet Generator, to assemble address-merged packets at runtime. Finally, a destination-aware SNN partitioning algorithm increases the overlap of post-synaptic destinations within each core, improving the opportunity for aggregation.

\par We evaluate UniSpike on two neuro-scientific SNN networks spanning three neuron models and six deep learning SNNs covering vision and NLP workloads. Compared with leading neuromorphic communication schemes, UniSpike achieves $1.93\times$ traffic savings, $1.77\times$ speedup, and $1.50\times$ energy efficiency improvement on average, outperforming state-of-the-art multicast and packet-compression approaches.

\par In summary, this paper makes three contributions. First, we characterize address redundancy in neuromorphic NoC traffic and show its significant impact across SNN workloads. Second, we introduce destination-centric spike scheduling and lightweight hardware support for efficient address-merged packet generation. Third, we develop destination-aware SNN partitioning to further maximize spike aggregation opportunities.

\section{Background and Motivation}\label{sec:background}
\subsection{Spiking Neural Networks}
\par SNNs can be viewed as directed graphs in which neurons communicate through discrete binary spikes \cite{neuromorphic-survey,SNN-survey2}. Each spike conveys only a firing event, so the payload carried by a spike packet is extremely small when SNNs are deployed on neuromorphic systems \cite{ma2024darwin3}.

\begin{figure}
    \centering
    \includegraphics[width=0.8\linewidth]{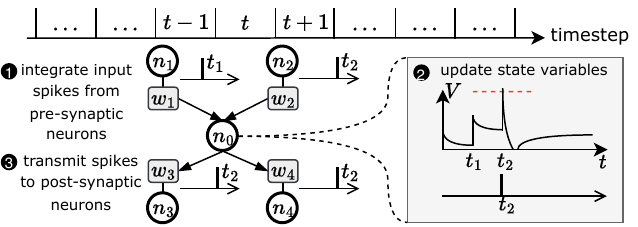}
    \caption{Structure and working mechanism of SNN.}
    \label{fig:background-snn}
\end{figure}

\par As shown in Figure \ref{fig:background-snn}, a neuron integrates spikes from pre-synaptic neurons and emits spikes to post-synaptic neurons when its state crosses a threshold. This computation proceeds over discrete timesteps, with neurons updating states and generating spikes in parallel.

\subsection{Neuromorphic Systems}
\par Neuromorphic systems \cite{debole2019truenorth,hoppner2021spinnaker2,davies2018loihi} organize homogeneous neuro-computing cores in a 2D mesh interconnected by an NoC. Each core stores synaptic weights, neuron states, and post-synaptic destinations in local SRAM, while system-level synchronization keeps cores processing the same logical timestep \cite{lee2022neurosync, davies2018loihi}.
\par Within each core, neurons are typically updated sequentially. Once a neuron fires, the core immediately sends spike packets to all destination cores hosting its post-synaptic neurons. We refer to this common design as \textit{neuron-centric} communication. Since post-synaptic weights are stored at the destination, each spike packet only needs to carry a minimal firing-neuron identifier, making address fields a disproportionately large part of each packet.

\subsection{SNN to Neuromorphic Systems Deployment}
\par In most compilers or toolchains for modern neuromorphic systems such as LCompiler \cite{lin2018mapping} for Loihi \cite{davies2018loihi}, PACMAN \cite{galluppi2012hierachical} for SpiNNaker \cite{hoppner2021spinnaker2}, NEUTRUM \cite{ji2016neutrams} for Tianji \cite{deng2020tianjic}, deploying an SNN onto many-core systems requires a two-stage process: (1) Partitioning, which divides the SNN into neuron clusters to fit core memory constraints, as a single core is insufficient to accommodate the entire network, and (2) Mapping, which assigns each cluster to a hardware core.

\subsection{Address Redundancy}\label{sec:motivation}
\par Communication overhead often dominates neuromorphic execution \cite{communication-bottleneck-loihi, communication-bottleneck-vision, communication-bound-tsmc}, with one study reporting a latency share of $89.9\%$ \cite{communication-bound-tsmc}. We find that the neuron-centric communication mechanism introduces substantial address redundancy that prior data/payload-oriented techniques do not directly address.
\par
\begin{figure}
    \centering
    \includegraphics[width=0.5\linewidth]{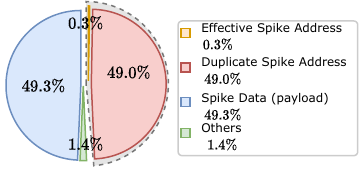}
    \caption{Profile of traffic distribution. The evaluation is performed on average across all workloads in Section \ref{sec:evaluation}.}
    % \vspace{-2em}
    \label{fig:moti-traffic}
\end{figure}

 We quantify this redundancy using the benchmarks in Section \ref{sec:evaluation}. For each source core and timestep, \textit{effective spike address} traffic counts only one address per unique destination, regardless of how many spikes are sent there. Figure \ref{fig:moti-traffic} shows that this effective address traffic can be less than $0.6\%$ of the total address traffic, revealing a large opportunity for address merging.

\section{Design of UniSpike}\label{sec:method}
\subsection{Overview}

\begin{figure}[!t]
    \centering
    \includegraphics[width=\linewidth]{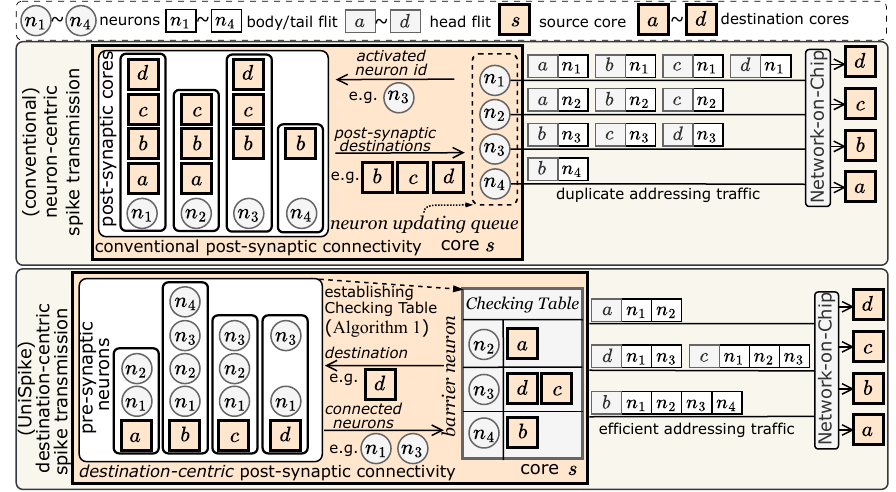}
    \caption{The comparison of working mechanisms between \textit{destination-centric} (proposed) and conventional \textit{neuron-centric} spike transmission scheduling.}
    \label{fig:comm-scheme}
\end{figure}

\par UniSpike reduces spike traffic by allowing packets with the same destination to share address data. It combines three components: destination-centric spike transmission scheduling, a hardware implementation that assembles address-merged packets at runtime, and destination-aware SNN partitioning that increases destination overlap among neurons placed in the same core.

\subsection{Spike Transmission Scheduling}\label{sec:transmission-scheduler}
\par Conventional neuron-centric communication sends packets immediately after each neuron fires. If later neurons share the same destination cores, the same addresses are transmitted again. UniSpike instead organizes dispatch from the destination perspective, scheduling spikes bound for the same core together as \textit{destination-centric} spike transmission.
\par For each core, UniSpike introduces \textit{barrier neurons}: synchronization points indicating that the activation states of all neurons connected to one or more destination cores have been determined. A compile-time \textit{Checking Table} maps barrier neurons to destination cores, allowing the runtime scheduler to aggregate and dispatch relevant spikes once each barrier is reached.
\begin{algorithm}[!t]
\small
\caption{Establishing Checking Table} \label{algo:checking-table}
\KwIn{map from post-synaptic destination to the connected neurons $P=\{(c_i:N_i)\}$}
\KwOut{neuron execution queue $Q$, checking table (map from barrier neuron to post-synaptic destinations) $M=\{(n_i:C_i)\}$}
$Q=[]$, $M=\{\}$; Sort $P$ in ascending order of $\lvert N_i \rvert$ \\
\For{$c_i$, $N_i$ in $P$} {
    $D \gets N_i \setminus Q$ \\
    \If{$D \neq []$} {
        Insert($Q, D$); Insert($M, (Q[-1]:c_i)$) \\
    } \Else {
        $n \gets $ the last neuron in $Q$ that belongs to $N_i$ \\
        \If{$n \notin M.keys$} {
            Insert($M, (n:c_i)$) \\
        } \Else {
            Insert($M[n], c_i$) \\
        }
    }    
}
\end{algorithm}
\par At each barrier neuron, the scheduler checks the activated neurons in the current timestep and collects, for each relevant destination, all activated pre-synaptic neuron identifiers connected to that destination. These identifiers are packed into one \textit{address-merged spike packet}, so a destination address is transmitted once for multiple spikes. Algorithm \ref{algo:checking-table} constructs the Checking Table by ordering destination cores and assigning the latest required neuron as the barrier for each destination.

\subsection{Implementation: UniSpike Architecture}\label{sec:architecture}
\par
\begin{figure}[t]
    \centering
    \includegraphics[width=\linewidth]{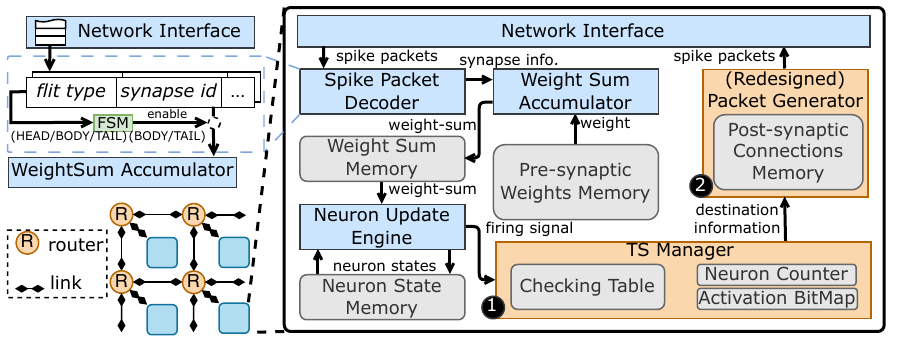}
    \caption{Overview of the UniSpike architecture.}
    \vspace{-18pt}
    \label{fig:micro-architecture}
\end{figure}

 \textbf{Overview}. UniSpike extends a conventional neuromorphic core \cite{davies2018loihi, lee2022neurosync, lee2022parallel} with two modules: a Transmission Scheduling (TS) Manager and a redesigned Packet Generator. Figure \ref{fig:micro-architecture} shows the overall architecture. The TS Manager triggers packet dispatch at barrier neurons, while the Packet Generator constructs address-merged packets.

\begin{figure}[t]
    \centering
    \includegraphics[width=0.85\linewidth]{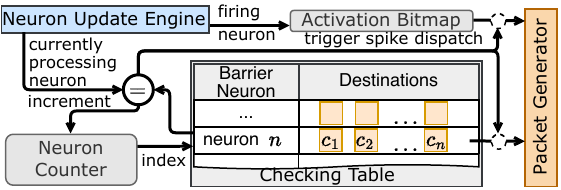}
    \caption{Working mechanism of TS Manager.}
    \vspace{-5pt}
    \label{fig:ts-manager-arch}
\end{figure}
\begin{figure}[t]
    \centering
    \includegraphics[width=0.85\linewidth]{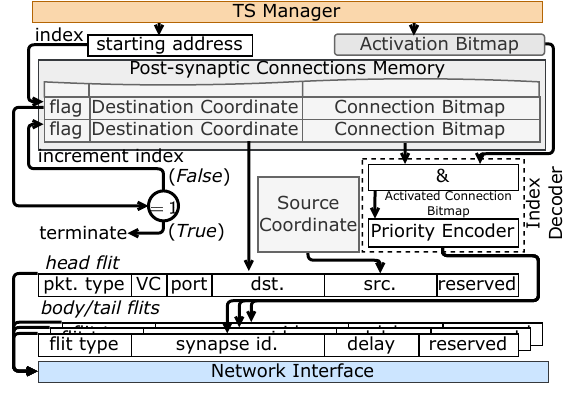}
    \caption{Working mechanism of Packet Generator.}
    \label{fig:pkt-generator-arch}
\end{figure}

\par \noindent \textbf{Baseline Core}. The baseline core contains a Network Interface, Spike Packet Decoder, Weight-Sum Accumulator, Neuron Update Engine, and Packet Generator. UniSpike keeps the computation datapath intact and only aggregates multiple payload flits into longer packets without changing the head/body flit formats. Therefore, the baseline decoder can process UniSpike packets without additional unpacking logic.

\par \noindent \textbf{TS Manager}. Figure \ref{fig:ts-manager-arch} illustrates the TS Manager. An Activation Bitmap records neurons that fire in the current timestep, and a Neuron Counter tracks the current barrier neuron. When the Neuron Update Engine completes a barrier neuron, the TS Manager triggers the Packet Generator for all destinations associated with that barrier. The Post-synaptic Connections Memory is ordered according to the Checking Table, so each destination entry serves as the starting address for packet generation.

\par \noindent \textbf{Redesigned Packet Generator}. Figure \ref{fig:pkt-generator-arch} shows the Packet Generator. It stores each destination coordinate and its connected local neurons as a Connection Bitmap. A bitwise AND between the Connection Bitmap and Activation Bitmap produces the activated neurons for that destination. The Index Decoder then emits these neuron identifiers as payload flits under a shared destination address until all active bits are consumed.

\subsection{Destination-Aware SNN Partitioning}\label{sec:snn-partitioning}
\par Destination-centric scheduling is most effective when neurons sharing post-synaptic destinations are placed in the same core. UniSpike therefore uses destination-aware partitioning to increase address-sharing opportunities during deployment. The algorithm has two phases: initial partitioning and swapping refinement.

\par \noindent \textbf{Initial Partitioning}. Partitioning is performed layer by layer. For convolutional layers, neurons at the same spatial position across channels share destinations, and nearby spatial positions often have overlapping destinations. UniSpike therefore orders neurons in the $WH$ plane using the Hilbert Space-Filling Curve (HSFC), preserving spatial locality while producing a one-dimensional sequence \cite{hilbert1935stetige}. The sequence is then segmented into clusters under core memory constraints.

\begin{figure}[!t]
    \centering
    \begin{subfigure}[b]{0.54\linewidth}
        \centering
        \includegraphics[width=\linewidth]{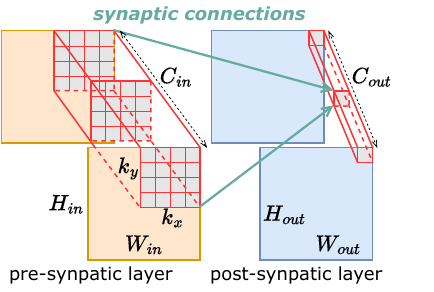}
        \caption{}
        \label{fig:conv}
    \end{subfigure}
    \hfill
    \begin{subfigure}[b]{0.324\linewidth}
        \centering
        \includegraphics[width=\linewidth]{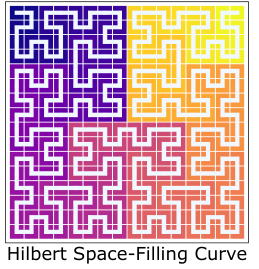}
        \caption{}
        \label{fig:hsc}
    \end{subfigure}
    \caption{(a) Synaptic connection pattern of convolution layers. (b) Hilbert Space-Filling Curve (HSFC). The locality-preserving property of HSFC provides an initial layout to aggregate neurons with common destinations together.}
    \label{fig:conv-hsc}
\end{figure}

\par \noindent \textbf{Swapping Refinement}. After initial partitioning, UniSpike applies Stochastic Segment Swap (SSS), inspired by simulated annealing, to exchange neuron segments between clusters. The objective is to reduce the number of distinct post-synaptic destinations inside each cluster, thereby increasing the probability that multiple spikes can share one destination address. The hyper-parameter $seg\_ratio$ controls the swapped segment size.

\section{Evaluation}\label{sec:evaluation}
\subsection{Experimental Setup}\label{sec:experiment-setup}
\subsubsection{Experimental Platform}
\par We implement a cycle-accurate simulator that tracks computation and data movement for latency and energy evaluation. The NoC is modeled using HeteroGarnet \cite{heterogarnet} with a $2$D mesh topology. A frontend based on SNNToolBox \cite{rueckauer2017conversion} performs SNN partitioning and generates per-core workloads for the simulator.

\begin{figure*}[!t]
    \centering
    \includegraphics[width=\linewidth]{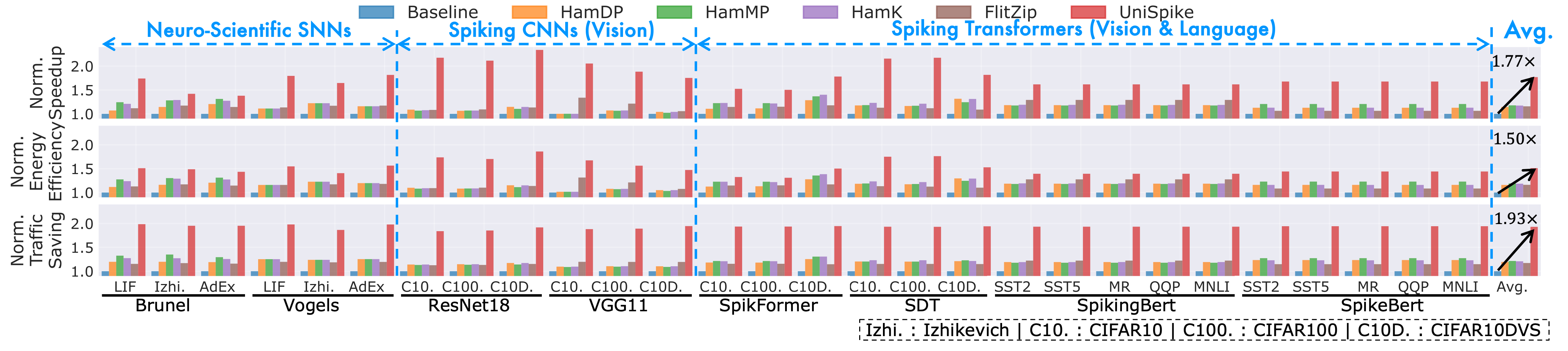}
    \caption{Speedup, Energy Efficiency improvement, and NoC Traffic Saving (normalized by \texttt{Baseline}) of \texttt{UniSpike}. The Neuro-Scientific SNNs differ in neuron models, while the Spiking CNNs and Spiking Transformers networks differ in datasets.}
    \vspace{-16pt}
    \label{fig:main-results}
\end{figure*}

\par Energy is modeled following standard methodologies \cite{chen2016eyeriss,kwon2019understanding}, using characterized memory-access and arithmetic-operation costs. Area overheads are evaluated with NeuroSim \cite{peng2020dnn+} at the $32$ nm CMOS technology node. Each core uses 100.75KB, 3KB, and 32.625KB SRAMs for synapses, neurons, and post-connections, respectively, plus 1.125KB for the Checking Table. As shown in Table \ref{tab:area}, UniSpike adds about $10.6\%$ area. Cores and NoC run at $500$ MHz and $160$ MHz, respectively; the NoC uses XY routing, 4 virtual channels, and 512 cores. We validate simulator functionality by comparing spike trains with Brian2 \cite{stimberg2019brian}; UniSpike is lossless and preserves accuracy.
\begin{figure}[t]
    \centering
    \includegraphics[width=0.75\linewidth]{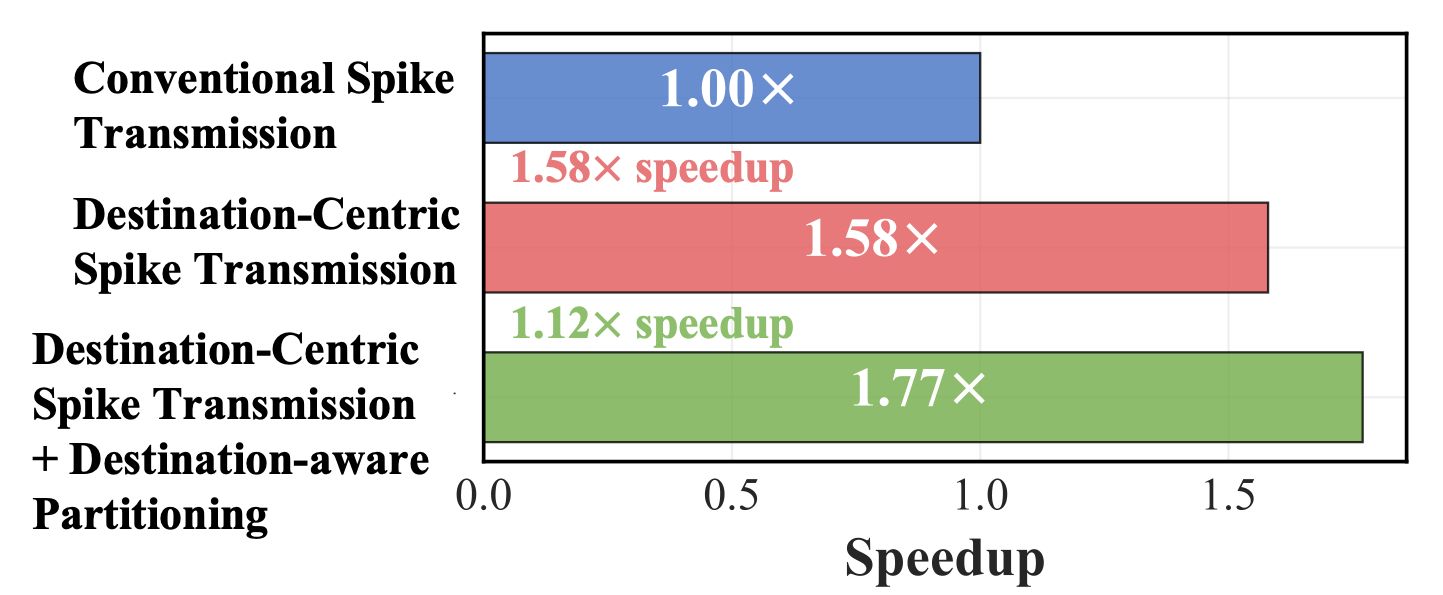}
    \caption{Ablation Study}
    \label{fig:ablation}
\end{figure}

\begin{table}[t]
    \centering
    \begin{tabularx}{\linewidth}{
    >{\centering\arraybackslash\hsize=0.4\hsize}X
    >{\centering\arraybackslash\hsize=0.7\hsize}X
    >{\centering\arraybackslash\hsize=0.6\hsize}X
    >{\centering\arraybackslash\hsize=1.4\hsize}X
    >{\centering\arraybackslash\hsize=0.8\hsize}X
    >{\centering\arraybackslash\hsize=1.6\hsize}X
    >{\centering\arraybackslash\hsize=0.5\hsize}X}
        \toprule
        Arch & \textbf{Syn.} & \textbf{Dec.} & \textbf{Che.-Tab.} & \textbf{Neuron} & \textbf{Post-Conn.} & \textbf{Total} \\
        \midrule
        \texttt{B} & 142.52 & 0.04 & - & 3.17 & 8.81 & 154.54 \\
        \texttt{U} & 142.52 & 0.04  & 0.87 & 3.17 & 24.35 & 170.95\\
        \bottomrule
    \end{tabularx}
    \caption{Area Cost ($mm^2$). \texttt{B} for \texttt{Baseline}. \texttt{U} for \texttt{UniSpike}.}
    \label{tab:area}
\end{table}

\subsubsection{Evaluated Workloads}
\par We benchmark UniSpike on neuro-scientific SNNs and modern spiking deep learning models. The neuro-scientific suite includes Brunel \cite{brunel2000dynamics} and Vogels \cite{vogels2011inhibitory} networks under LIF \cite{LIF}, Izhikevich \cite{izhikevich2003simple}, and AdEx \cite{adex} neuron models. The deep learning suite includes spiking CNNs (ResNet18 \cite{he2016deep}, VGG11 \cite{simonyan2014very}) and spiking Transformers (SpikFormer \cite{zhou2022spikformer}, SDT \cite{yao2023spike-driven-transformer}, SpikingBERT \cite{bal2024spikingbert}, and SpikeBert \cite{lv2023spikebert}) across vision and NLP datasets \cite{CIFAR,CIFAR10DVS,SST2-SST5,MR,QQP,MNLI}.

\subsubsection{Methods for Comparison}
\par We compare UniSpike against conventional spike transmission (\texttt{Baseline}) \cite{davies2018loihi, ma2024darwin3}, multicast-based methods \texttt{HamDP} \cite{HamDP}, \texttt{HamMP} \cite{HamMP}, and \texttt{HamK} \cite{HamK}, and the packet compression technique \texttt{FlitZip} \cite{deb2021flitzip}. All methods are implemented in the same simulation framework with identical configurations.

\subsection{Overall Results}
\par \noindent \textbf{Traffic Saving}. Figure \ref{fig:main-results} shows that UniSpike achieves $1.93\times$ traffic savings over \texttt{Baseline} while delivering the same spikes. The reduction comes from replacing repeated destination-address transmissions with address-merged packets.
\par \noindent \textbf{Speedup}. UniSpike delivers an average $1.77\times$ speedup by reducing communication latency. It outperforms \texttt{FlitZip} because spike packets have minimal payload and little intra-packet structure to compress. It also outperforms multicast methods by merging spikes to a common destination regardless of their neuronal origin.
\par \noindent \textbf{Energy Efficiency}. UniSpike improves energy efficiency by $1.50\times$. Shorter execution reduces static energy, while fewer addressing flits reduce dynamic activity in routers, links, and packet unpacking logic.

\subsection{Ablation Study}
\par Figure \ref{fig:ablation} reports averaged ablation results. Destination-centric scheduling alone achieves $1.58\times$ speedup by eliminating redundant address traffic. Adding destination-aware SNN partitioning increases destination overlap within each core, improving spike merging opportunities. With both components enabled, UniSpike reaches $1.77\times$ average speedup over the baseline.

\section{Conclusion}\label{sec:conclusion}
\par This work presents \texttt{UniSpike}, a hardware-software co-designed communication scheme that targets overlooked address redundancy in neuromorphic systems. By integrating destination-centric spike scheduling, lightweight hardware support, and destination-aware SNN partitioning, \texttt{UniSpike} eliminates duplicate address traffic by merging spikes with common destinations. Evaluation shows this approach achieves $1.93\times$ traffic reduction, $1.77\times$ speedup, and $1.50\times$ energy efficiency improvement on average. This work offers a path toward efficient and sustainable neuromorphic computing.

\bibliographystyle{ACM-Reference-Format}
\bibliography{final_sample-base}

@article{communication-bound-tsmc,
  title={MLG-NCS: Multimodal local--global neuromorphic computing system for affective video content analysis},
  author={Ji, Xiaoyue and Dong, Zhekang and Zhou, Guangdong and Lai, Chun Sing and Qi, Donglian},
  journal={IEEE Trans. Syst., Man, Cybern.: Syst.},
  volume={54},
  number={8},
  pages={5137--5149},
  year={2024}
}

@article{communication-bottleneck-vision,
  title={Neuromorphic computing for robotic vision: algorithms to hardware advances},
  author={Chowdhury, Sayeed Shafayet and Sharma, Deepika and Kosta, Adarsh and Roy, Kaushik},
  journal={Commun. Eng.},
  volume={4},
  number={1},
  pages={152},
  year={2025}
}

@article{communication-bottleneck-loihi,
  title={Advancing neuromorphic computing with loihi: A survey of results and outlook},
  author={Davies, Mike and Wild, Andreas and Orchard, Garrick and Sandamirskaya, Yulia and Guerra, Gabriel A Fonseca and Joshi, Prasad and Plank, Philipp and Risbud, Sumedh R},
  journal={Proc. IEEE},
  volume={109},
  number={5},
  pages={911--934},
  year={2021}
}

@article{SNN-survey2,
  author = {GHOSH-DASTIDAR, SAMANWOY and ADELI, HOJJAT},
  title = {SPIKING NEURAL NETWORKS},
  journal = {Intl. J. Neural Syst.},
  volume = {19},
  number = {04},
  pages = {295-308},
  year = {2009},
  doi = {10.1142/S0129065709002002}
}

@inproceedings{communication-bottleneck2,
  title={Deep spiking neural network: Energy efficiency through time based coding},
  author={Han, Bing and Roy, Kaushik},
  booktitle={Proc. Eur. Conf. Comput. Vis. (ECCV)},
  pages={388--404},
  year={2020}
}

@article{neuromorphic-survey,
  title={A survey of neuromorphic computing and neural networks in hardware},
  author={Schuman, Catherine D and Potok, Thomas E and Patton, Robert M and Birdwell, J Douglas and Dean, Mark E and Rose, Garrett S and Plank, James S},
  journal={arXiv},
  eprint = {1705.06963},
  year={2017}
}

@ARTICLE{multicast-path-TC14,
  author={Ebrahimi, Masoumeh and Daneshtalab, Masoud and Liljeberg, Pasi and Plosila, Juha and Flich, José and Tenhunen, Hannu},
  journal={IEEE Trans. Comput.}, 
  title={Path-Based Partitioning Methods for 3D Networks-on-Chip with Minimal Adaptive Routing}, 
  year={2014},
  volume={63},
  number={3},
  pages={718-733},
  doi={10.1109/TC.2012.255}
}

@incollection{hilbert1935stetige,
  title={{\"U}ber die stetige Abbildung einer Linie auf ein Fl{\"a}chenst{\"u}ck},
  author={Hilbert, David},
  booktitle={Dritter Band: Analysis...},
  pages={1--2},
  year={1935},
  publisher={Springer}
}

@InProceedings{MNLI,
  author = "Williams, Adina and Nangia, Nikita and Bowman, Samuel",
  title = "A Broad-Coverage Challenge Corpus for Sentence Understanding through Inference",
  booktitle = {Proc. 2018 Conf. NAACL: HLT, Vol. 1},
  year = "2018",
  pages = "1112--1122",
  url = "http://aclweb.org/anthology/N18-1101"
}

@article{QQP,
  title={GLUE: A multi-task benchmark and analysis platform for natural language understanding},
  author={Wang, Alex and Singh, Amanpreet and Michael, Julian and Hill, Felix and Levy, Omer and Bowman, Samuel R},
  journal={arXiv},
  eprint = {1804.07461},
  year={2018}
}

@inproceedings{MR,
   author = {Bo Pang and Lillian Lee},
   title = {Seeing stars: Exploiting class relationships for sentiment categorization with respect to rating scales},
   year = {2005},
   pages = {115--124},
   booktitle = {Proc. ACL}
}

@inproceedings{SST2-SST5,
    title = "Recursive Deep Models for Semantic Compositionality Over a Sentiment Treebank",
    author = "Socher, Richard  and Perelygin, Alex  and Wu, Jean  and Chuang, Jason  and Manning, Christopher D.  and Ng, Andrew  and Potts, Christopher",
    booktitle = {Proc. 2013 Conf. Empirical Methods Natural Lang. Process. (EMNLP)},
    year = "2013",
    pages = "1631--1642",
}

@article{zhou2022spikformer,
  title={Spikformer: When spiking neural network meets transformer},
  author={Zhou, Zhaokun and Zhu, Yuesheng and He, Chao and Wang, Yaowei and Yan, Shuicheng and Tian, Yonghong and Yuan, Li},
  journal={arXiv},
  eprint = {2209.15425},
  year={2022}
}

@article{yao2023spike-driven-transformer,
  title={Spike-driven transformer},
  author={Yao, Man and Hu, Jiakui and Zhou, Zhaokun and Yuan, Li and Tian, Yonghong and Xu, Bo and Li, Guoqi},
  journal={Proc. Adv. Neural Inf. Process. Syst. (NeurIPS)},
  volume={36},
  pages={64043--64058},
  year={2023}
}

@article{lv2023spikebert,
  title={Spikebert: A language spikformer learned from bert with knowledge distillation},
  author={Lv, Changze and Li, Tianlong and Xu, Jianhan and Gu, Chenxi and Ling, Zixuan and Zhang, Cenyuan and Zheng, Xiaoqing and Huang, Xuanjing},
  journal={arXiv},
  eprint = {2308.15122},
  year={2023}
}

@inproceedings{bal2024spikingbert,
  title={Spikingbert: Distilling bert to train spiking language models using implicit differentiation},
  author={Bal, Malyaban and Sengupta, Abhronil},
  booktitle={Proc. AAAI Conf. Artif. Intell. (AAAI)},
  volume={38},
  number={10},
  pages={10998--11006},
  year={2024}
}

@article{simonyan2014very,
  title={Very deep convolutional networks for large-scale image recognition},
  author={Simonyan, Karen and Zisserman, Andrew},
  journal={arXiv},
  eprint = {1409.1556},
  year={2014}
}

@inproceedings{he2016deep,
  title={Deep residual learning for image recognition},
  author={He, Kaiming and Zhang, Xiangyu and Ren, Shaoqing and Sun, Jian},
  booktitle={Proc. IEEE Conf. Comput. Vis. Pattern Recogn. (CVPR)},
  pages={770--778},
  year={2016}
}

@article{vogels2011inhibitory,
  title={Inhibitory plasticity balances excitation and inhibition in sensory pathways and memory networks},
  author={Vogels, Tim P and Sprekeler, Henning and Zenke, Friedemann and Clopath, Claudia and Gerstner, Wulfram},
  journal={Science},
  volume={334},
  number={6062},
  pages={1569--1573},
  year={2011}
}

@article{lin2018mapping,
  title={Mapping spiking neural networks onto a manycore neuromorphic architecture},
  author={Lin, Chit-Kwan and Wild, Andreas and Chinya, Gautham N and Lin, Tsung-Han and Davies, Mike and Wang, Hong},
  journal={ACM SIGPLAN Notices},
  volume={53},
  number={4},
  pages={78--89},
  year={2018}
}

@article{ma2024darwin3,
  title={Darwin3: a large-scale neuromorphic chip with a novel ISA and on-chip learning},
  author={Ma, De and Jin, Xiaofei and Sun, Shichun and Li, Yitao and Wu, Xundong and Hu, Youneng and Yang, Fangchao and Tang, Huajin and Zhu, Xiaolei and Lin, Peng and others},
  journal={Natl. Sci. Rev.},
  volume={11},
  number={5},
  pages={nwae102},
  year={2024}
}

@article{deng2020tianjic,
  title={Tianjic: A unified and scalable chip bridging spike-based and continuous neural computation},
  author={Deng, Lei and Wang, Guanrui and Li, Guoqi and Li, Shuangchen and Liang, Ling and Zhu, Maohua and Wu, Yujie and Yang, Zheyu and Zou, Zhe and Pei, Jing and others},
  journal={IEEE J. Solid-State Circuits (JSSC)},
  volume={55},
  number={8},
  pages={2228--2246},
  year={2020}
}

@inproceedings{ji2016neutrams,
  title={NEUTRAMS: Neural network transformation and co-design under neuromorphic hardware constraints},
  author={Ji, Yu and Zhang, YouHui and Li, ShuangChen and Chi, Ping and Jiang, CiHang and Qu, Peng and Xie, Yuan and Chen, WenGuang},
  booktitle={Proc. 49th Annu. IEEE/ACM Intl. Symp. Microarch. (MICRO)},
  pages={1--13},
  year={2016}
}

@inproceedings{galluppi2012hierachical,
  title={A hierachical configuration system for a massively parallel neural hardware platform},
  author={Galluppi, Francesco and Davies, Sergio and Rast, Alexander and Sharp, Thomas and Plana, Luis A and Furber, Steve},
  booktitle={Proc. 9th Conf. Comput. Frontiers (CF)},
  pages={183--192},
  year={2012}
}

@inproceedings{jin2008adaptive,
  title={Adaptive data compression for high-performance low-power on-chip networks},
  author={Jin, Yuho and Yum, Ki Hwan and Kim, Eun Jung},
  booktitle={Proc. 41st IEEE/ACM Intl. Symp. Microarch. (MICRO)},
  pages={354--363},
  year={2008}
}

@inproceedings{das2008performance,
  title={Performance and power optimization through data compression in network-on-chip architectures},
  author={Das, Reetuparna and Mishra, Asit K and Nicopoulos, Chrysostomos and Park, Dongkook and Narayanan, Vijaykrishnan and Iyer, Ravishankar and Yousif, Mazin S and Das, Chita R},
  booktitle={Proc. 14th IEEE Intl. Symp. High Perform. Comput. Arch. (HPCA)},
  pages={215--225},
  year={2008}
}

@article{hoppner2021spinnaker2,
  title={The SpiNNaker 2 processing element architecture for hybrid digital neuromorphic computing},
  author={H{\"o}ppner, Sebastian and Yan, Yexin and Dixius, Andreas and Scholze, Stefan and Partzsch, Johannes and Stolba, Marco and Kelber, Florian and Vogginger, Bernhard and Neum{\"a}rker, Felix and Ellguth, Georg and others},
  journal={arXiv},
  eprint = {2103.08392},
  year={2021}
}

@article{debole2019truenorth,
  title={TrueNorth: Accelerating from zero to 64 million neurons in 10 years},
  author={DeBole, Michael V and Taba, Brian and Amir, Arnon and Akopyan, Filipp and Andreopoulos, Alexander and Risk, William P and Kusnitz, Jeff and Otero, Carlos Ortega and Nayak, Tapan K and Appuswamy, Rathinakumar and others},
  journal={IEEE Comput.},
  volume={52},
  number={5},
  pages={20--29},
  year={2019}
}

@inproceedings{HamMP,
  title={An efficent dynamic multicast routing protocol for distributing traffic in NOCs},
  author={Ebrahimi, Masoumeh and Daneshtalab, Masoud and Neishaburi, Mohammad Hossein and Mohammadi, Siamak and Afzali-Kusha, Ali and Plosila, Juha and Tenhunen, Hannu},
  booktitle={Proc. Design, Autom. Test Eur. Conf. Exhib. (DATE)},
  pages={1064--1069},
  year={2009}
}

@article{HamDP,
  title={Multicast communication in multicomputer networks},
  author={Lin, Xiaola and Ni, Lionel M},
  journal={IEEE Trans. Parallel Distrib. Syst.},
  volume={4},
  number={10},
  pages={1105--1117},
  year={2002}
}

@article{HamK,
  title={Path-based multicast routing for network-on-chip of the neuromorphic processor},
  author={Kang, Zi-Yang and Li, Shi-Ming and Wang, Shi-Ying and Qu, Lian-Hua and Gong, Rui and Shi, Wei and Xu, Wei-Xia and Wang, Lei},
  journal={J. Comput. Sci. Technol.},
  volume={38},
  number={5},
  pages={1098--1112},
  year={2023}
}

@article{deb2021flitzip,
  title={Flitzip: Effective packet compression for noc in multiprocessor system-on-chip},
  author={Deb, Dipika and Rohith, MK and Jose, John},
  journal={IEEE Trans. Parallel Distrib. Syst.},
  volume={33},
  number={1},
  pages={117--128},
  year={2021}
}

@article{chen2016eyeriss,
  title={Eyeriss: An energy-efficient reconfigurable accelerator for deep convolutional neural networks},
  author={Chen, Yu-Hsin and Krishna, Tushar and Emer, Joel S and Sze, Vivienne},
  journal={IEEE J. Solid-State Circuits (JSSC)},
  volume={52},
  number={1},
  pages={127--138},
  year={2016}
}

@techreport{CIFAR,
  title={Learning multiple layers of features from tiny images},
  author={Krizhevsky, Alex and Hinton, Geoffrey},
  year={2009},
  institution={University of Toronto}
}

@article{CIFAR10DVS,
  title={Cifar10-dvs: an event-stream dataset for object classification},
  author={Li, Hongmin and Liu, Hanchao and Ji, Xiangyang and Li, Guoqi and Shi, Luping},
  journal={Front. Neurosci.},
  volume={11},
  pages={309},
  year={2017}
}

@article{izhikevich2003simple,
  title={Simple model of spiking neurons},
  author={Izhikevich, Eugene M},
  journal={IEEE Trans. Neural Netw.},
  volume={14},
  number={6},
  pages={1569--1572},
  year={2003}
}

@article{adex,
  title={Adaptive exponential integrate-and-fire model as an effective description of neuronal activity},
  author={Brette, Romain and Gerstner, Wulfram},
  journal={J. Neurophysiol.},
  volume={94},
  number={5},
  pages={3637--3642},
  year={2005}
}

@article{LIF,
  title={A review of the integrate-and-fire neuron model: I. Homogeneous synaptic input},
  author={Burkitt, Anthony N},
  journal={Biol. Cybern.},
  volume={95},
  pages={1--19},
  year={2006}
}

@article{brunel2000dynamics,
  title={Dynamics of sparsely connected networks of excitatory and inhibitory spiking neurons},
  author={Brunel, Nicolas},
  journal={J. Comput. Neurosci.},
  volume={8},
  pages={183--208},
  year={2000}
}

@article{stimberg2019brian,
  title={Brian 2, an intuitive and efficient neural simulator},
  author={Stimberg, Marcel and Brette, Romain and Goodman, Dan FM},
  journal={elife},
  volume={8},
  pages={e47314},
  year={2019}
}

@article{rueckauer2017conversion,
  title={Conversion of continuous-valued deep networks to efficient event-driven networks for image classification},
  author={Rueckauer, Bodo and Lungu, Iulia-Alexandra and Hu, Yuhuang and Pfeiffer, Michael and Liu, Shih-Chii},
  journal={Front. Neurosci.},
  volume={11},
  pages={682},
  year={2017}
}

@article{peng2020dnn+,
  title={DNN+ NeuroSim V2. 0: An end-to-end benchmarking framework for compute-in-memory accelerators for on-chip training},
  author={Peng, Xiaochen and Huang, Shanshi and Jiang, Hongwu and Lu, Anni and Yu, Shimeng},
  journal={IEEE Trans. Comput.-Aided Design Integr. Circuits Syst. (TCAD)},
  volume={40},
  number={11},
  pages={2306--2319},
  year={2020}
}

@inproceedings{kwon2019understanding,
  title={Understanding reuse, performance, and hardware cost of dnn dataflow: A data-centric approach},
  author={Kwon, Hyoukjun and Chatarasi, Prasanth and Pellauer, Michael and Parashar, Angshuman and Sarkar, Vivek and Krishna, Tushar},
  booktitle={Proc. 52nd Annu. IEEE/ACM Intl. Symp. Microarch. (MICRO)},
  pages={754--768},
  year={2019}
}

@inproceedings{heterogarnet,
    author={Bharadwaj, Srikant and Yin, Jieming and Beckmann, Bradford and Krishna, Tushar},
    booktitle={Proc. 57th ACM/IEEE Design Autom. Conf. (DAC)},
    title={Kite: A Family of Heterogeneous Interposer Topologies Enabled via Accurate Interconnect Modeling},
    year={2020},
    pages={1-6},
    doi={10.1109/DAC18072.2020.9218539}
}

@article{davies2018loihi,
  title={Loihi: A neuromorphic manycore processor with on-chip learning},
  author={Davies, Mike and Srinivasa, Narayan and Lin, Tsung-Han and Chinya, Gautham and Cao, Yongqiang and Choday, Sri Harsha and Dimou, Georgios and Joshi, Prasad and Imam, Nabil and Jain, Shweta and others},
  journal={IEEE Micro},
  volume={38},
  number={1},
  pages={82--99},
  year={2018}
}

@inproceedings{lee2022parallel,
  title={Parallel time batching: Systolic-array acceleration of sparse spiking neural computation},
  author={Lee, Jeong-Jun and Zhang, Wenrui and Li, Peng},
  booktitle={Proc. 2022 IEEE Intl. Symp. High-Perform. Comput. Arch. (HPCA)},
  pages={317--330},
  year={2022}
}

@inproceedings{lee2022neurosync,
  title={Neurosync: A scalable and accurate brain simulator using safe and efficient speculation},
  author={Lee, Hunjun and Kim, Chanmyeong and Kim, Minseop and Chung, Yujin and Kim, Jangwoo},
  booktitle={Proc. 2022 IEEE Intl. Symp. High-Perform. Comput. Arch. (HPCA)},
  pages={633--647},
  year={2022}
}

\end{document}